\documentclass[a4paper,english,journal,onecolumn,draftclsnofoot,12pt]{IEEEtran}

\renewcommand{\baselinestretch}{1.4}

\IEEEoverridecommandlockouts

\usepackage{gensymb}
\usepackage{dsfont}
\usepackage{amsmath,amssymb,amsfonts,amsthm}

\usepackage{epsfig}
\usepackage{cite}
\usepackage{hhline}
\usepackage{multirow}
\usepackage{xcolor}
\usepackage{makecell}
\usepackage{subcaption}
\usepackage{capt-of}
\usepackage[labelformat=simple]{subcaption}

\usepackage{enumerate}
\usepackage{enumitem}
\usepackage{color}
\usepackage{graphicx}
\usepackage{algpseudocode}
\usepackage[ruled]{algorithm}
\makeatletter
\newcounter{parentalgorithm}

\makeatother

\usepackage[nolist,printonlyused]{acronym}      
\usepackage{booktabs}
\usepackage{bm}
\usepackage{pgf}
\usepackage{pgfplots}
\usepackage{tikz}
\usepackage{grffile}
\pgfplotsset{compat=newest}
\usetikzlibrary{plotmarks}
\usetikzlibrary{shapes.geometric,quotes,angles,automata,arrows,positioning,calc,3d,matrix,decorations.markings,matrix}

\tikzstyle{startstop} = [rectangle, rounded corners, minimum width=3cm, minimum height=1cm,text centered, draw=black, fill=white!30]
\tikzstyle{io} = [trapezium, trapezium left angle=70, trapezium right angle=110, minimum width=3cm, minimum height=1cm, text centered, draw=black, fill=white!30]
\tikzstyle{process} = [rectangle, minimum width=2cm, minimum height=1cm, text centered, draw=black, fill=white!30]
\tikzstyle{decision} = [diamond, minimum width=3cm, minimum height=1cm, text centered, draw=black, fill=white!30]
\tikzstyle{arrow} = [thick,->,>=stealth]

\makeatletter
\newcommand{\vast}{\bBigg@{3}}
\newcommand{\Vast}{\bBigg@{4}}
\makeatother
\renewcommand{\IEEEQED}{\IEEEQEDopen}

\showoutput
\makeatletter
\tikzset{%
  remember picture with id/.style={%
    remember picture,
    overlay,
    save picture id=#1,
  },
  save picture id/.code={%
    \edef\pgf@temp{#1}%
    \immediate\write\pgfutil@auxout{%
      \noexpand\savepointas{\pgf@temp}{\pgfpictureid}}%
  },
  if picture id/.code args={#1#2#3}{%
    \@ifundefined{save@pt@#1}{%
      \pgfkeysalso{#3}%
    }{
      \pgfkeysalso{#2}%
    }
  }
}

\def\savepointas#1#2{%
  \expandafter\gdef\csname save@pt@#1\endcsname{#2}%
}

\def\tmk@labeldef#1,#2\@nil{%
  \def\tmk@label{#1}%
  \def\tmk@def{#2}%
}

\tikzdeclarecoordinatesystem{pic}{%
  \pgfutil@in@,{#1}%
  \ifpgfutil@in@%
    \tmk@labeldef#1\@nil
  \else
    \tmk@labeldef#1,(0pt,0pt)\@nil
  \fi
  \@ifundefined{save@pt@\tmk@label}{%
    \tikz@scan@one@point\pgfutil@firstofone\tmk@def
  }{%
  \pgfsys@getposition{\csname save@pt@\tmk@label\endcsname}\save@orig@pic%
  \pgfsys@getposition{\pgfpictureid}\save@this@pic%
  \pgf@process{\pgfpointorigin\save@this@pic}%
  \pgf@xa=\pgf@x
  \pgf@ya=\pgf@y
  \pgf@process{\pgfpointorigin\save@orig@pic}%
  \advance\pgf@x by -\pgf@xa
  \advance\pgf@y by -\pgf@ya
  }%
}

\makeatother

\usepackage{comment}
\usepackage{xpatch}
\makeatletter
\xpatchcmd{\algorithmic}{\itemsep\z@}{\itemsep=-0.25mm}{}{}
\makeatother

\usepackage{matlab-prettifier}

\usepackage{multicol}
\usepackage{multirow}

\usepackage{lipsum}
\usepackage{multicol}

\usepackage{amsmath}
\usepackage{algorithm}

\algrenewcommand\algorithmicforall{\textbf{foreach}}
\algrenewcommand\algorithmicindent{.8em}
\usepackage{pgfplots}
\usetikzlibrary{pgfplots.groupplots}
\begin{document}

\title{A-LAQ: Adaptive Lazily 

Aggregated Quantized Gradient}

\author{Afsaneh Mahmoudi$^{1}$, José Mairton Barros Da Silva Júnior$^{1,2}$, \\Hossein S. Ghadikolaei$^{3}$, and Carlo Fischione$^{1}$
\\ $^{1}$Network and Systems Engineering, \\$^{1}$Electrical Engineering and Computer Science \\$^{1}$KTH Royal Institute of Technology, Stockholm, Sweden \\
$^{1}$emails: \{afmb, jmbdsj, carlofi\}@kth.se \\$^2$ Princeton University, NJ, USA
\\$^3$ Ericsson, Stockholm, Sweden\\
$^3$\{email: hossein.shokri.ghadikolaei@ericsson.com\}} 


\newtheorem{theorem}{Theorem}
\newtheorem{defin}{Definition}
\newtheorem{prop}{Proposition}
\newtheorem{lemma}{Lemma}
\newtheorem{corollary}{Corollary}
\newtheorem{alg}{Algorithm}
\newtheorem{remark}{Remark}
\newtheorem{result}{Result}
\newtheorem{conjecture}{Conjecture}
\newtheorem{example}{Example}
\newtheorem{notations}{Notations}
\newtheorem{assumption}{Assumption}
\newcommand{\combin}[2]{\ensuremath{ \left( \ba{c} #1 \\ #2 \ea \right) }}
\newcommand{\diag}{{\mbox{diag}}}
\newcommand{\rank}{{\mbox{rank}}}
\newcommand{\dom}{{\mbox{dom{\color{white!100!black}.}}}}
\newcommand{\range}{{\mbox{range{\color{white!100!black}.}}}}
\newcommand{\image}{{\mbox{image{\color{white!100!black}.}}}}
\newcommand{\herm}{^{\mbox{\scriptsize H}}}  
\newcommand{\sherm}{^{\mbox{\tiny H}}}       
\newcommand{\tran}{^{\mbox{\scriptsize T}}}  
\newcommand{\tranIn}{^{\mbox{-\scriptsize T}}}  
\newcommand{\card}{{\mbox{\textbf{card}}}}
\newcommand{\asign}{{\mbox{$\colon\hspace{-2mm}=\hspace{1mm}$}}}
\newcommand{\ssum}[1]{\mathop{ \textstyle{\sum}}_{#1}}

\newcommand{\vbar}{\raisebox{.17ex}{\rule{.04em}{1.35ex}}}
\newcommand{\vbarind}{\raisebox{.01ex}{\rule{.04em}{1.1ex}}}
\newcommand{\D}{\ifmmode {\rm I}\hspace{-.2em}{\rm D} \else ${\rm I}\hspace{-.2em}{\rm D}$ \fi}
\newcommand{\T}{\ifmmode {\rm I}\hspace{-.2em}{\rm T} \else ${\rm I}\hspace{-.2em}{\rm T}$ \fi}
\newcommand{\B}{\ifmmode {\rm I}\hspace{-.2em}{\rm B} \else \mbox{${\rm I}\hspace{-.2em}{\rm B}$} \fi}
\newcommand{\Hil}{\ifmmode {\rm I}\hspace{-.2em}{\rm H} \else \mbox{${\rm I}\hspace{-.2em}{\rm H}$} \fi}
\newcommand{\C}{\ifmmode \hspace{.2em}\vbar\hspace{-.31em}{\rm C} \else \mbox{$\hspace{.2em}\vbar\hspace{-.31em}{\rm C}$} \fi}
\newcommand{\Cind}{\ifmmode \hspace{.2em}\vbarind\hspace{-.25em}{\rm C} \else \mbox{$\hspace{.2em}\vbarind\hspace{-.25em}{\rm C}$} \fi}
\newcommand{\Q}{\ifmmode \hspace{.2em}\vbar\hspace{-.31em}{\rm Q} \else \mbox{$\hspace{.2em}\vbar\hspace{-.31em}{\rm Q}$} \fi}
\newcommand{\Z}{\ifmmode {\rm Z}\hspace{-.28em}{\rm Z} \else ${\rm Z}\hspace{-.38em}{\rm Z}$ \fi}

\newcommand{\sgn}{\mbox {sgn}}
\newcommand{\var}{\mbox {var}}
\newcommand{\E}{\mbox {E}}
\newcommand{\cov}{\mbox {cov}}
\renewcommand{\Re}{\mbox {Re}}
\renewcommand{\Im}{\mbox {Im}}
\newcommand{\cum}{\mbox {cum}}

\renewcommand{\vec}[1]{{\bf{#1}}}     

\newcommand{\vecsc}[1]{\mbox {\boldmath \scriptsize $#1$}}     
\newcommand{\itvec}[1]{\mbox {\boldmath $#1$}}
\newcommand{\itvecsc}[1]{\mbox {\boldmath $\scriptstyle #1$}}
\newcommand{\gvec}[1]{\mbox{\boldmath $#1$}}

\newcommand{\balpha}{\mbox {\boldmath $\alpha$}}
\newcommand{\bbeta}{\mbox {\boldmath $\beta$}}
\newcommand{\bgamma}{\mbox {\boldmath $\gamma$}}
\newcommand{\bdelta}{\mbox {\boldmath $\delta$}}
\newcommand{\bepsilon}{\mbox {\boldmath $\epsilon$}}
\newcommand{\bvarepsilon}{\mbox {\boldmath $\varepsilon$}}
\newcommand{\bzeta}{\mbox {\boldmath $\zeta$}}
\newcommand{\boldeta}{\mbox {\boldmath $\eta$}}
\newcommand{\btheta}{\mbox {\boldmath $\theta$}}
\newcommand{\bvartheta}{\mbox {\boldmath $\vartheta$}}
\newcommand{\biota}{\mbox {\boldmath $\iota$}}
\newcommand{\blambda}{\mbox {\boldmath $\lambda$}}
\newcommand{\bmu}{\mbox {\boldmath $\mu$}}
\newcommand{\bnu}{\mbox {\boldmath $\nu$}}
\newcommand{\bxi}{\mbox {\boldmath $\xi$}}
\newcommand{\bpi}{\mbox {\boldmath $\pi$}}
\newcommand{\bvarpi}{\mbox {\boldmath $\varpi$}}
\newcommand{\brho}{\mbox {\boldmath $\rho$}}
\newcommand{\bvarrho}{\mbox {\boldmath $\varrho$}}
\newcommand{\bsigma}{\mbox {\boldmath $\sigma$}}
\newcommand{\bvarsigma}{\mbox {\boldmath $\varsigma$}}
\newcommand{\btau}{\mbox {\boldmath $\tau$}}
\newcommand{\bupsilon}{\mbox {\boldmath $\upsilon$}}
\newcommand{\bphi}{\mbox {\boldmath $\phi$}}
\newcommand{\bvarphi}{\mbox {\boldmath $\varphi$}}
\newcommand{\bchi}{\mbox {\boldmath $\chi$}}
\newcommand{\bpsi}{\mbox {\boldmath $\psi$}}
\newcommand{\bomega}{\mbox {\boldmath $\omega$}}

\newcommand{\R}{\mathbb{R}}
\newcommand{\N}{\mathbb{N}}

\def\calA{{\mathcal A}}
\def\calB{{\mathcal B}}
\def\calC{{\mathcal C}}
\def\calD{{\mathcal D}}
\def\calE{{\mathcal E}}
\def\calF{{\mathcal F}}
\def\calG{{\mathcal G}}
\def\calH{{\mathcal H}}
\def\calI{{\mathcal I}}
\def\calJ{{\mathcal J}}
\def\calK{{\mathcal K}}
\def\calL{{\mathcal L}}
\def\calM{{\mathcal M}}
\def\calN{{\mathcal N}}
\def\calO{{\mathcal O}}
\def\calP{{\mathcal P}}
\def\calQ{{\mathcal Q}}
\def\calR{{\mathcal R}}
\def\calS{{\mathcal S}}
\def\calT{{\mathcal T}}
\def\calU{{\mathcal U}}
\def\calV{{\mathcal V}}
\def\calW{{\mathcal W}}
\def\calX{{\mathcal X}}
\def\calY{{\mathcal Y}}
\def\calZ{{\mathcal Z}}

\def\bA{\mbox {\boldmath $A$}}
\def\bB{\mbox {\boldmath $B$}}
\def\bC{\mbox {\boldmath $C$}}
\def\bD{\mbox {\boldmath $D$}}
\def\bE{\mbox {\boldmath $E$}}
\def\bF{\mbox {\boldmath $F$}}
\def\bG{\mbox {\boldmath $G$}}
\def\bH{\mbox {\boldmath $H$}}
\def\bI{\mbox {\boldmath $I$}}
\def\bJ{\mbox {\boldmath $J$}}
\def\bK{\mbox {\boldmath $K$}}
\def\bL{\mbox {\boldmath $L$}}
\def\bM{\mbox {\boldmath $M$}}
\def\bN{\mbox {\boldmath $N$}}
\def\bO{\mbox {\boldmath $O$}}
\def\bP{\mbox {\boldmath $P$}}
\def\bQ{\mbox {\boldmath $Q$}}
\def\bR{\mbox {\boldmath $R$}}
\def\bS{\mbox {\boldmath $S$}}
\def\bT{\mbox {\boldmath $T$}}
\def\bU{\mbox {\boldmath $U$}}
\def\bV{\mbox {\boldmath $V$}}
\def\bW{\mbox {\boldmath $W$}}
\def\bX{\mbox {\boldmath $X$}}
\def\bY{\mbox {\boldmath $Y$}}
\def\bZ{\mbox {\boldmath $Z$}}

\def\ba{\mbox {$\bf{a}$}}
\def\bb{\mbox {\boldmath $b$}}
\def\bc{\mbox {\boldmath $c$}}
\def\bd{\mbox {\boldmath $d$}}
\def\be{\mbox {\boldmath $e$}}
\def\bg{\mbox {\boldmath $g$}}
\def\bh{\mbox {\boldmath $h$}}
\def\bi{\mbox {\boldmath $i$}}
\def\bj{\mbox {\boldmath $j$}}
\def\bk{\mbox {\boldmath $k$}}
\def\bl{\mbox {\boldmath $l$}}
\def\bm{\mbox {\boldmath $m$}}
\def\bn{\mbox {\boldmath $n$}}
\def\bo{\mbox {\boldmath $o$}}
\def\bp{\mbox {\boldmath $p$}}
\def\bq{\mbox {\boldmath $q$}}
\def\br{\mbox {\boldmath $r$}}
\def\bs{\mbox {\boldmath $s$}}
\def\bt{\mbox {\boldmath $t$}}
\def\bu{\mbox {\boldmath $u$}}
\def\bv{\mbox {\boldmath $v$}}
\def\bw{\mbox {\boldmath $w$}}
\def\bx{\mbox {\boldmath $x$}}
\def\by{\mbox {\boldmath $y$}}
\def\bz{\mbox {\boldmath $z$}}

\newcommand{\snr}{\textup{SNR}}
\newcommand{\UE}{\mathrm{UE}}
\newcommand{\BS}{\mathrm{BS}}
\newcommand{\Passoc}{p_{_{I^{(1)}}}}
\newcommand{\Pintra}{p_{_{I^{(2)}}}}
\newcommand{\Pinter}{p_{_{I^{(3)}}}}

\newenvironment{Ex}
{\begin{adjustwidth}{0.04\linewidth}{0cm}
\begingroup\small
\vspace{-1.0em}
\raisebox{-.2em}{\rule{\linewidth}{0.3pt}}
\begin{example}
}
{
\end{example}
\vspace{-5mm}
\rule{\linewidth}{0.3pt}
\endgroup
\end{adjustwidth}}

\newcommand{\Hossein}[1]{{\textcolor{blue}{\emph{**Hossein: #1**}}}}
\newcommand{\Gabor}[1]{{\textcolor{cyan}{\emph{**Afsaneh: #1**}}}}
\newcommand{\Hadi}[1]{{\textcolor{red}{#1}}}
\newcommand{\gf}[1]{{\textcolor{cyan}{#1}}}
\newcommand{\REV}[1]{{\textcolor{blue}{#1}}}


\makeatletter
\let\save@mathaccent\mathaccent
\newcommand*\if@single[3]{%
  \setbox0\hbox{${\mathaccent"0362{#1}}^H$}%
  \setbox2\hbox{${\mathaccent"0362{\kern0pt#1}}^H$}%
  \ifdim\ht0=\ht2 #3\else #2\fi
  }
\newcommand*\rel@kern[1]{\kern#1\dimexpr\macc@kerna}
\newcommand*\widebar[1]{\@ifnextchar^{{\wide@bar{#1}{0}}}{\wide@bar{#1}{1}}}
\newcommand*\wide@bar[2]{\if@single{#1}{\wide@bar@{#1}{#2}{1}}{\wide@bar@{#1}{#2}{2}}}
\newcommand*\wide@bar@[3]{%
  \begingroup
  \def\mathaccent##1##2{%
    \let\mathaccent\save@mathaccent
    \if#32 \let\macc@nucleus\first@char \fi
    \setbox\z@\hbox{$\macc@style{\macc@nucleus}_{}$}%
    \setbox\tw@\hbox{$\macc@style{\macc@nucleus}{}_{}$}%
    \dimen@\wd\tw@
    \advance\dimen@-\wd\z@
    \divide\dimen@ 3
    \@tempdima\wd\tw@
    \advance\@tempdima-\scriptspace
    \divide\@tempdima 10
    \advance\dimen@-\@tempdima
    \ifdim\dimen@>\z@ \dimen@0pt\fi
    \rel@kern{0.6}\kern-\dimen@
    \if#31
      \overline{\rel@kern{-0.6}\kern\dimen@\macc@nucleus\rel@kern{0.4}\kern\dimen@}%
      \advance\dimen@0.4\dimexpr\macc@kerna
      \let\final@kern#2%
      \ifdim\dimen@<\z@ \let\final@kern1\fi
      \if\final@kern1 \kern-\dimen@\fi
    \else
      \overline{\rel@kern{-0.6}\kern\dimen@#1}%
    \fi
  }%
  \macc@depth\@ne
  \let\math@bgroup\@empty \let\math@egroup\macc@set@skewchar
  \mathsurround\z@ \frozen@everymath{\mathgroup\macc@group\relax}%
  \macc@set@skewchar\relax
  \let\mathaccentV\macc@nested@a
  \if#31
    \macc@nested@a\relax111{#1}%
  \else
    \def\gobble@till@marker##1\endmarker{}%
    \futurelet\first@char\gobble@till@marker#1\endmarker
    \ifcat\noexpand\first@char A\else
      \def\first@char{}%
    \fi
    \macc@nested@a\relax111{\first@char}%
  \fi
  \endgroup
}
\makeatother

\def\herm{\mathsf{H}}
\def\trans{\mathsf{T}}
\newcommand{\call}[1]{{\textsf{\small \textsc{#1}}}}
\newcommand{\callf}[1]{{\textsf{\footnotesize \textsc{#1}}}}

\def\argmax{\mathrm{arg}\max}
\def\argmin{\mathrm{arg}\min}
\renewcommand{\algorithmicrequire}{\textbf{Input:}}
\renewcommand{\algorithmicensure}{\textbf{Output:}}
\algdef{SE}[PROCEDURE]{Procedure}{EndProcedure}%
   [2]{\algorithmicprocedure\ \textproc{#1}\ifthenelse{\equal{#2}{}}{}{(#2)}}%
   {\algorithmicend\ \algorithmicprocedure}%
\algdef{SE}[FUNCTION]{Function}{EndFunction}%
   [2]{\algorithmicfunction\ \textproc{#1}\ifthenelse{\equal{#2}{}}{}{(#2)}}%
   {\algorithmicend\ \algorithmicfunction}%

\makeatletter
\newcommand\fs@betterruled{%
  \def\@fs@cfont{\bfseries}\let\@fs@capt\floatc@ruled
  \def\@fs@pre{\vspace*{5pt}\hrule height.8pt depth0pt \kern2pt}%
  \def\@fs@post{\kern2pt\hrule\relax}%
  \def\@fs@mid{\kern2pt\hrule\kern2pt}%
  \let\@fs@iftopcapt\iftrue}
\floatstyle{betterruled}
\restylefloat{algorithm}
\makeatother

\maketitle
\vspace{-10mm}
\begin{abstract}
Federated Learning~(FL) plays a prominent role in solving machine learning problems with data distributed across clients. In FL, to reduce the communication overhead of data between clients and the server, each client communicates the local FL parameters instead of the local data. However, when a wireless network connects clients and the server, the communication resource limitations of the clients may prevent completing the training of the FL iterations. Therefore, communication-efficient variants of FL have been widely investigated. Lazily Aggregated Quantized Gradient~(LAQ) is one of the promising communication-efficient approaches to lower resource usage in FL. However, LAQ assigns a fixed number of bits for all iterations, which may be communication-inefficient when the number of iterations is medium to high or convergence is approaching. This paper proposes Adaptive Lazily Aggregated Quantized Gradient~(A-LAQ), which is a method that significantly extends LAQ by assigning an adaptive number of communication bits during the FL iterations. We train FL in an energy-constraint condition and investigate the convergence analysis for A-LAQ. The experimental results highlight that A-LAQ outperforms LAQ by up to a $50$\% reduction in spent communication energy and an $11$\% increase in test accuracy.
\end{abstract}
\section{Introduction}

Federated Learning~(FL) is a framework in which the clients train a centralized model by communicating their computed local models while data remains at each client~\cite{konevcny2016federated}. FL has been widely studied because it preserves local data privacy and reduces communication overhead by avoiding data transmission. FL clients contribute to FL training by computing and sharing a local FL vector. However, computation and communication of such local vectors in large-scale FL require extensive communication resources~\cite{hellstrom2022wireless}. Furthermore, the resources needed for FL training may be available in wired networks but not on wireless devices due to communication and energy resource constraints. Thus, we must minimize communication resource expenditure and get the most accurate training possible.  

Many papers have recently focused on communication, computation, latency, and energy-efficient FL~\cite{9264742, 8917592, mahmoudi2020cost,mahmoudi20spawc,mahmoudi2022fedcau}. Authors in~\cite{9264742} have tried to minimize the system's total spent communication energy under a latency constraint and could reduce up to 59.5~\% energy expenditure compared to the conventional FL. Reference~\cite{8917592} studied the joint power and resource allocation for ultra-reliable low-latency communication in vehicular networks and proposed a distributed approach based on FL to estimate the tail distribution of the queue lengths. Finally, authors of~\cite{mahmoudi2020cost,mahmoudi20spawc,mahmoudi2022fedcau} have proposed a causal setting to jointly minimize the FL loss function and the overall resource consumption for training. Their results highlighted that joint design of communication protocols and FL are crucial for resource-efficient and accurate FL training.

Besides resource optimization, communication-efficient methods like quantization~\cite{amiri2020federated,9305988}, compression~\cite{8889996}, and sparsification~\cite{9148987} can significantly reduce the communication overhead at each communication iteration. Adaptive methods have been recently noticed for communication-efficient FL training~\cite{ 9413697,  9207963,  9355797, 9728013}. Authors in~\cite{9413697} have proposed an adaptive quantization strategy named AdaQuantFL by which they can change the quantization level in the stochastic quantization method to improve communication efficiency. Reference~\cite{9207963} has considered an adaptive quantization and sparsification scheme for uplink transmission facilitated by non-orthogonal multiple access. Authors in~\cite{9355797} have proposed an online learning scheme for determining the communication and computation trade-off. This trade-off is controlled by the degree of gradient sparsity obtained by the estimated sign of the objective function's derivative. Authors of~\cite{9728013} have proposed an adaptive gradient compression approach that improves communication efficiency by adjusting the compression rate according to the actual characteristics of each client.

Lazily aggregated quantized gradients~(LAQ) method~\cite{9238427 } is a novel framework that achieves the same linear convergence as the gradient descent in strongly convex set-ups. In addition, LAQ saves communication resources by using fewer transmitted bits at each communication iteration. However, LAQ considers a constant number of bits at each global and local FL transmission, which may not be communication-efficient enough.

In this paper, we significantly extend LAQ by considering an adaptive number of bits during the FL training to further improve communication and resource efficiency. 
The critical factors in our proposed method are the descent behavior and the \textit{diminishing return} rule~\cite{9563954} in FL training for $L$-smooth and convex loss functions. Due to the diminishing return rule, the accuracy improvement of the final model reduces with every new local and global communication iteration. Thus, we propose an adaptive LAQ, which we called A-LAQ, in which the FL training starts with a higher number of communication bits and adapts the bits as the communication between the server and clients continues. As the number of communication iterations increases, we propose that the number of bits can either decrease or stay the same. In A-LAQ, we assign more communication bits to the first communication iterations to minimize the quantization error at the beginning steps of training. After some communication iterations, we reduce the number of communication bits while facing a minor reduction in the loss function during training. We also develop a convergence analysis of FL with A-LAQ. The numerical results show that energy-constraint FL with A-LAQ outperforms FL with LAQ by up to a $50$\% reduction in spent communication energy and an $11$\% increase in test accuracy.

 We organize the rest of this paper as the following. Section~\ref{section: System_Model} describes the general system model and problem formulation. In Section~\ref{section:Solution Approach}, we explain the solution approaches and convergence analysis for A-LAQ. Section~\ref{section: numerical} shows some numerical results of A-LAQ and its performance compared to LAQ, and we conclude the paper in Section~\ref{section: conclusion}.

\emph{Notation:} Normal font $w$, bold font small-case $\bw$, bold-font capital letter $\bW$, and calligraphic font $\calW$ denote scalar, vector, matrix, and set, respectively. We define the index set $[N] = \{1,2,\ldots,N\}$ for any integer $N$. We denote by $\|\cdot\|$ the $l_2$-norm, by $\lceil. \rceil$ the ceiling value, by $|\calA|$ the cardinality of set $\calA$, by $[\bw]_i$ the entry $i$ of vector $\bw$, by $\bw{\tran}$ the transpose of $\bw$, and $\mathds{1}_{x}$ is an indicator function taking $1$ if and only if $x$ is true and takes $0$ otherwise.


\section{System Model and Problem Formulation}\label{section: System_Model}

In this section, we represent the system model and the problem formulation. 
Consider a star network of $M$ worker nodes that cooperatively solve a distributed training problem involving a loss function $f(\bw)$. Consider $D$ as the whole dataset distributed among each worker $j \in [M]$ with $D_j$ data samples. Let tuple $(\bx_{ij}, y_{ij})$ denote data sample $i$ of $|D_j|$ samples of worker node $j$ and $\bw \in \R^d$ denote the model parameter at the master node. Considering $\sum_{j =1}^{ M} |D_j| = |D|$, and $j, j' \in [M]$, $j \neq j'$, we assume $D_j \cap D_{j'}=\emptyset$, and defining $\rho^j := {|D_j|}/{|D|}$, we formulate the following training problem
\vspace{-0.005\textheight}
\begin{equation}\label{eq: our-optimization}
\bw^* \in \argmin_{\bw\in\R^d} f(\bw)=\sum_{j =1}^{ M} 
{\rho^j f^j(\bw)},
\end{equation}
where $f^j(\bw) := \sum_{i =1}^{|D_j|} {f(\bw; \bx_{ij}, y_{ij})}/{|D_j|}$. 

\subsection{LAQ Summary}
In this part, we briefly summarize LAQ and its important parameters~\cite{9238427}. Considering the communication bits $b$, we define the quantization granularity $\tau := 1/(2^b-1)$, the quantized version of each local gradient at the global communication iteration~$k$ as~$\bq^j(\bw_k) =  {\text{Quant}}(\nabla f^j(\bw_k);b), j \in [M]$. Each local gradient is element-wise quantized by projecting to the closest point in a uniformly discretized $d$-dimensional grid with radius of~$R_k^j= \| \nabla f^j(\bw_k) - \bq^j({\bw}_{k-1})\|_\infty$. We assume that all the workers participate in the training, each local loss function $f^j(\bw_{k})$ is $L_j$-smooth, the aggregated loss function $f(\bw_k)$ is $L$-smooth and $\mu$-strongly convex. Defining $\varepsilon_k^j:= \nabla f^j(\bw_k) - \bq^j(\bw_k)$ as the local quantization error, the aggregated quantization error is obtained as $\varepsilon_k:= \sum_{j=1}^M \varepsilon_k^j$ and the aggregated quantized gradient is $\bq_k := \sum_{j=1}^M \bq^j(\bw_k)$.
 The global updates in LAQ is $\bw_{k} = \bw_{k-1} - \alpha \Tilde{\nabla}_{k-1}$, where $\Tilde{\nabla}_k = \Tilde{\nabla}_{k-1} + \sum_{j=1}^{M} \delta \bq_k^j$ and $\delta \bq_k^j := \bq^j(\bw_k) - \bq^j(\bw_{k-1})$.
 \subsection{Adaptive LAQ}\label{subsec: adaptive laq} 

In this subsection, we propose A-LAQ, in which we let $b_k$ be the adaptive number of communication bits, and we introduce $\tau_k:=1/(2^{b_k}-1)$ at each communication iteration $1\le k \le K$. The global update in FL with A-LAQ is similar to LAQ, but the number of communication bits $b_k$ becomes adaptive. First, we propose the following optimization problem, which formalizes the general scope of this paper:
\begin{subequations}\label{eq: general}
\begin{alignat}{3}
\label{genral1}
  \underset{\bw, k_0, K}{\mathrm{minimize}} & \quad f(\bw) \: \\ 
  \text{subject to}& 
 \label{general2}
 \quad \bw_{k} = \bw_{k-1} - \alpha \Tilde{\nabla}_{k-1},\quad k=1,\ldots, K \: \\
 \label{general21}
&\quad \Tilde{\nabla}_k = \Tilde{\nabla}_{k-1} + \sum_{j=1}^{M} \delta \bq_k^j,\quad k=1,\ldots, K \: \\
 \label{general0}
& \quad \delta \bq_k^j = \bq^j(\bw_{k}) - \bq^j(\bw_{k-1}) , \quad k=1,\ldots, K \: \\
\label{generl1}
&\quad b_k = b^{\max} \mathds{1}_{k\le k_0}  \: \\
\nonumber
& \qquad +b_0 \mathds{1}_{k=k_0+1}+ \lceil \eta_{k-1}b_{k-1}\rceil\mathds{1}_{k>k_0+1}\: \\
\label{cost bmin}
 &\quad b_k \geq 2,\quad k=1,\ldots, K \: \\
\label{E1}
&\quad \sum_{k=1}^K E_k \le E, \quad k=1,\ldots,K,
\end{alignat}
\end{subequations}
where $\bw_k$ is the global FL parameter at each communication iteration $k$, $\rho^j, j \in [M]$ is the local weight, $\alpha$ is the step size, $E_k$ is the communication energy spent at each communication iteration~$k$, $E$ is the total communication energy budget, and $k_0 \le K$ is the number of the first communication iterations by which we assign $b_k = b^{\max}$, where $b^{\max}$ and $b_0$ are the given number of bits. 
We propose to update $b_{k} = \lceil \eta_{k-1}b_{k-1}\rceil$ for$~k= \max\{3,k_0\},\ldots, K$, by introducing~$\eta_{k-1}$ as
\begin{equation}\label{eq: eta}
    \eta_{k-1} := \min \left\{ \frac{\|f(\bw_{k-1})- f(\bw_{k-2}) \|}{\|f(\bw_{k-2})- f(\bw_{k-3}) \|}, 1\right\},
\end{equation}
where the rationale of such a choice is the diminishing return rule. Constraints~\eqref{general2}-\eqref{general0} reveal global LAQ update, constraints~\eqref{generl1}~and~\eqref{cost bmin} show the adaptive $b_k$, and constraint~\eqref{E1} is the overall communication energy limitation. 

Optimization problem~\eqref{eq: general} aims to solve an FL problem in a communication energy-limited set-up. Although LAQ is a promising communication-efficient method, we show that under the same resource limitation, A-LAQ saves more communication resources than LAQ. The set-up for A-LAQ is to assign a high number of communication bits to the communication iterations $1, \ldots,k_0$. Afterward, the training continues with $b_0$ communication bits, while $b_0 < b$ (where recall that $b$ is the number of bits used by LAQ), and follows a non-increasing sequence of bits as implied by~\eqref{eq: eta}. 

Optimization problem~\eqref{eq: general} is not practical because it requires $K$ and the future local gradients for $k=1,\ldots, K$ at the beginning of the training. Since it is impossible to have the information of local parameters and $K$ beforehand, we call such a problem~\textit{non-causal}~\cite{mahmoudi2020cost}. Therefore, in the rest of this paper, we focus on developing causal and practical solution approaches which do not need the future information of local gradients and $K$.
\section{Solution Approach}\label{section:Solution Approach}
This section provides a solution approach for optimization problem~\eqref{eq: general}. Since optimization problem~\eqref{eq: general} is non-causal, we first calculate $k_0$, then proceed to calculate $K$ and $\bw^*$ in a causal way. To obtain $k_0$, we propose to solve a new optimization problem demonstrating the effect of the diminishing return rule on energy expenditure. After computing $k_0$, we simplify the optimization problem~\eqref{eq: general} and solve it to find $K$ and $\bw^*$ causally until the energy budget constraint is fulfilled.


\subsection{Preliminary Results}\label{subsec: Preliminary Results}
To calculate $k_0$, we propose an optimization problem considering the diminishing return rule and energy expenditure. The idea behind A-LAQ is to change the number of communication bits to cope with the diminishing return rule. In other words, A-LAQ tries to associate a different number of communication bits at each communication iteration~$k$ to save the extra communication energy the clients spend before FL converges. Therefore, we define the energy-per-progress ratio function $ E_f(\bw_k,k; \calI_k^j)$, where $\calI_k^j$ is set of network's clients parameters, as 
\begin{equation}\label{eq: g}
    E_f(\bw_k,k; M, [p_{k}^j]_{j}, [t_{k}^j]_{j}) := \frac{\sum_{k'=1}^{k} \sum_{j=1}^{M} p_{k'}^j t_{k'}^j}{f(\bw_0)-f(\bw_k)},~k\ge 1,
\end{equation}
where~$p_{k'}^j$ and~$t_{k'}^j$ are respectively the transmission power and latency of each client~$j \in [M]$ at every communication iteration~$k'= 1,\ldots,k$. We assume that the client powers are constant at each communication iteration~$k'$, as $p_{k'}^j = p^j, j\in[M]$. Defining client transmission rate~$r^j$ bits/sec, we compute the transmission latency for each client $j\in[M]$, as~$t_{k'}^j=b_{k'}d/r^j$~sec, where $d$ is the dimension of the local and global parameters. Consider $r^j$ as
\begin{equation}\label{eq: r^j}
    r^j = \text{BW}^j \log_2 \left( 1 + \frac{p^j H^j}{N_0 \text{BW}^j} \right), 
\end{equation}
where $N_0$ is the power spectrum density of noise, $H^j$ is the channel gain and $\text{BW}^j$ is the bandwidth allocated to each client $j\in[M]$. Defining power vector $\bp :=[p^1, \ldots, p^M]$, bit vector~$\bb:=[b_1,\ldots,b_K]$, and the rate vector $\br :=[r^1,\ldots,r^M]$, we have
\begin{alignat}{3}\label{eq: Ef}
  E_f(\bw_k, k;& \bb, M, \bp, \br ) = \frac{\sum_{k'=1}^{k} E_{k'}}{f(\bw_0)-f(\bw_k)} = \: \\
  \nonumber
  &\frac{\sum_{k'=1}^{k} b_{k'}\sum_{j=1}^{M} \frac{p^j d}{\text{BW}^j \log_2 ( 1 + \frac{p^j H^j}{N_0 })}}{f(\bw_0)-f(\bw_k)},~k=1,\ldots, K.
\end{alignat}
Now, considering~$\bb = b^{\max} {\boldsymbol{{\textbf{1}}}}$, we aim to minimize~$E_f(\bw_k, k; \bb, M, \bp, \br )$ as
\begin{subequations}\label{eq: k0}
\begin{alignat}{3}
\label{k0_1}
  \underset{k, \bw, K}{\mathrm{minimize}} & \quad E_f(\bw_k, k; b^{\max}{\boldsymbol{{\textbf{1}}}}, M, \bp, \br ) \: \\ 
  \text{subject to}& 
 \label{geral2}
 \quad \bw_{k} = \bw_{k-1} - \alpha \Tilde{\nabla}_{k-1},\quad k=1,\ldots, K \: \\
 \label{ral21}
&\quad \Tilde{\nabla}_k = \Tilde{\nabla}_{k-1} + \sum_{j=1}^{M} \delta \bq_k^j,\quad k=1,\ldots, K \: \\
 \label{genel0}
& \quad \delta \bq_k^j = \bq^j(\bw_{k}) - \bq^j(\bw_{k-1}) , \quad k=1,\ldots, K \: \\
 \label{genel01}
& \quad f(\bw_k)=\sum_{j =1}^{ M} 
{\rho^j f^j(\bw_k)}, \quad k=1,\ldots, K, \: \\
\label{genel1}
& \quad \sum_{k=1}^K E_k \le E.
\end{alignat}
\end{subequations}

To solve optimization problem~\eqref{eq: k0}, we propose the following Lemma, which demonstrates the conditions for discrete convexity~\cite{miller1971minimizing} of $E_f(\bw_k, k; b^{\max}, M, \bp, \br )$.
\begin{lemma}\label{lemma: g}
Let $f(\bw)$ be $\mu$-strongly convex and $L$-smooth. Assume $b^{\max} = 32$ bits which represents the quantization full accuracy. Then, $E_f(\bw_k, k; b^{\max}, M, \bp, \br )$ is discrete convex w.r.t. $k$.
\end{lemma}
\begin{IEEEproof}
See Appendix~\ref{p: lemma: g}
\end{IEEEproof}
Lemma~\ref{lemma: g} demonstrates that $E_f(\bw_k, k; b^{\max}, M, \bp, \br )$ has a unique minimum w.r.t. $k$. Thus, we calculate $k_0$ as

\begin{subequations}\label{eq: cost-efficient-distributed-optim-3}
\begin{alignat}{3}
\label{general13}
 k_0 &\in \underset{k \in \N}{\argmin}  \hspace{1.6mm} E_f(\bw_k, k; b^{\max}, M, \bp, \br ) \: \\ 
 \text{subject to}& 
\label{general23}
 \quad \eqref{geral2}-\eqref{genel1}.
 \end{alignat}
 \end{subequations}
After computing~$k_0$, we re-write the optimization problem~\eqref{eq: general} as
\begin{subequations}\label{eq: short general}
\begin{alignat}{3}
\label{general1}
  \underset{\bw, K, \bb}{\mathrm{minimize}} & \quad f(\bw) \: \\ 
  \text{subject to}& 
  \label{sgeneral1}
\quad b_k = b_0 \mathds{1}_{k=k_0+1} + \lceil \eta_{k-1}b_{k-1}\rceil\mathds{1}_{k>k_0+1} \: \\
\label{sgeneral bmin}
 &\quad b_k \geq 2, \quad k=k_0+1,\ldots,K \: \\
\label{E11}
&\quad \sum_{k=k_0+1}^K E_k \le E - \sum_{k=1}^{k_0} E_k \: \\
\label{sgeneral23}
& \quad \eqref{general2}-\eqref{general0}.
\end{alignat}
\end{subequations}

Now, equipped with the preliminary results of this subsection, we are ready to solve optimization problem~\eqref{eq: general} in the following subsection.

\subsection{Solution Approach}\label{subsec: solution approach}

First, we consider Lemma~\ref{lemma: g} and compute $k_0$ according to the following proposition.
\begin{prop}\label{prop: k_0}
Let $f(\bw)$ be $\mu$-strongly convex and $L$-smooth. Consider $b^{\max} = 32$ bits. Thus, $k_0 = \min\{ k_e ,k_f\}$, where
\begin{equation}\label{eq: ke}
    k_e := {\textrm{the first value of}}~k~{\textrm{such that}}~E_k > E-\sum_{k'=1}^{k-1} E_{k'},
\end{equation}
and
\begin{equation}\label{eq: kf}
    k_f := {\textrm{the first value of}}~k~{\textrm{such that}} \qquad \quad
\end{equation}
$$
\qquad  k < \frac{ f(\bw_0)-f(\bw_k)}{f(\bw_{k-1})-f(\bw_k)}.
$$
\end{prop}
\begin{IEEEproof}
See Appendix~\ref{p: prop: k_0}
\end{IEEEproof}

Note that when $k_0 = k_e$, constraint~\eqref{E1} is fulfilled, thus the training is complete and $K=k_0$, $\bb = b^{\max}{\boldsymbol{{\textbf{1}}}}$. Otherwise, after computing $k_0$, we focus on optimization problem~\eqref{eq: short general} to obtain $K$, $\bw$ and $\bb$. Considering the non-increasing sequence of $b_k$ for $k~\in~\left[k_0+1,K\right]$ in~\eqref{sgeneral1} and~\eqref{sgeneral bmin} along with the energy constraint of~\eqref{E11}, we obtain
\begin{equation}\label{eq: Eb}
    \left(32 k_0 + b_0+\sum_{k'=k_0+2}^{K} b_{k'}\right)\sum_{j=1}^{M} \frac{p^j d }{\text{BW}^j \log_2 ( 1 + \frac{p^j H^j}{N_0 })} \le E.
\end{equation}
Eq.~\eqref{eq: Eb} plays a critical role in FL training for the communication iteration $k \ge k_0 +1$. It means that $K$ is obtained while the energy budget $E$ is spent. The following lemma determines when we can terminate the FL with A-LAQ training by finding $K$.
\begin{lemma}\label{lemma: EK}
Let $f(\bw)$ be $\mu$-strongly convex and $L$-smooth and $b^{\max} = 32$ bits. For any $k > k_0$, we obtain $K = k$ if

\begin{equation}\label{eq: EK}
  \eta_k b_k\sum_{j=1}^{M}  \frac{p^j d}{\text{BW}^j \log_2 ( 1 + \frac{p^j H^j}{N_0 })} > E - \sum_{k'=1}^{k} E_k.
\end{equation}
\end{lemma}
\begin{IEEEproof}
See Appendix~\ref{p: lemma: EK}
\end{IEEEproof}
Therefore, the FL training with A-LAQ continues until $K$ is obtained. Algorithm~\ref{alg: A-LAQ FL} summarizes all the steps for FL with A-LAQ.
\begin{algorithm}[t]
\caption{Federated Learning with A-LAQ}
\small
\label{alg: A-LAQ FL}
\begin{algorithmic}[1]
\State \textbf{Inputs:} $\bw_0$, $M$, ${(\bx_{ij}, y_{ij})}_{i,j}$, $\alpha$, $b^{\max}$, $b_0$, $\br$, $\bp$, $\{|D_j|\}_{j\in[M]}$, $\{\rho^j\}_{j\in[M]}$, $\mu$, $L$.
\State \textbf{Initialize:} $\Tilde{\nabla}_0$, $K = +\infty$, $k_0 =k_e = k_f = 0$, $(b_k)_{k\in[K]} = b^{\max}$
\State Master node broadcasts $\bw_0$ to all nodes
\While{$K = +\infty$}
\For{$k=1,\ldots, K$}
\For{$j \in [M]$} 

\State Calculate $\nabla f^j (\bw_k)$, $\bq^j (\bw_k)$, $\delta \bq_k^j$ and $f^j(\bw_k)$  

\State Send $\delta \bq_k^j$ and $f^j(\bw_k)$ to the master node
\EndFor 

\State Wait until master node collects all $\{\delta \bq_k^j\}_{j\in[M]}$ and update $f(\bw_{k})$, and $\Tilde{\nabla}_k$ and $\bw_{k}$ according to~\eqref{general2},~\eqref{general21}

\If{$k_0=0$} 
\If{$\max \{k_f, k_e \}> 0$}
\State Set $k_0 = k$
\State Set $b_{k+1} = b_0$
\EndIf 
\Else 
\State Calculate $\eta_k$ according to~\eqref{eq: eta}
\State Set $b_{k+1} = \lceil\eta_k b_k\rceil$
\If{Inequality~\eqref{eq: EK} is true} 
\State Set K = k
\EndIf 
\EndIf 
\State Set $k \leftarrow k+1 $
\EndFor
\EndWhile
\State \textbf{Return} $\bw_K$, $k_0$, $K$, $(b_k)_{k \in [K]}$
\end{algorithmic}
\end{algorithm}

\begin{theorem}\label{theorem: exact solution}
Let $f(\bw)$ be $\mu$-strongly convex and $L$-smooth. Assume $b^{\max} = 32$ and $b_0 < b$ be given. Then, by solving optimization problems~\eqref{eq: k0} and~\eqref{eq: short general}, we achieve an exact solution for optimization problem~\eqref{eq: general}.
\end{theorem}
\begin{IEEEproof}
In this paper, we propose to solve optimization problem~\eqref{eq: general} in a causal way. Thus, we first have to compute $k_0$ to determine when we must adapt the number of bits. To do so, we propose to solve optimization problem~\eqref{eq: k0} which highlights the diminishing return rule and energy expenditure. The solution to~\eqref{eq: k0} is exact and mathematically calculated by either~\eqref{eq: ke} or~\eqref{eq: kf}. Next, calculate $K$ and $\bw$, which is another causal approach, and the exact solution for $K$ is obtained by~\eqref{eq: EK}. 
\end{IEEEproof}

\subsection{Convergence Analysis}\label{subsec: convergence}
In this subsection, we investigate the convergence of A-LAQ. 
Since $\|\varepsilon_{k}^j \|_\infty \le \tau_k R_{k}^{j}$, for each element of $[\varepsilon_k^j]_i, i=1,\ldots,d$, we have $|[\varepsilon_k^j]_i| \le \tau_k R_k^j$, thus
\begin{equation}\label{eq: eps}
\|\varepsilon_k^j\|_2 \le \sqrt{d} \hspace{0.5mm} \tau_k R_k^j.
\end{equation}
According to definition of $\varepsilon_k$ in LAQ, $\varepsilon_k = \sum_{j=1}^{M} \varepsilon_k^j$, thus
\begin{equation}\label{eq: abs_epsk}
\|\varepsilon_k\|_2 = \left \|\sum_{j=1}^{M} \varepsilon_k^j \right\|_2  \stackrel{\small \text{ triangle}}{\le} \sum_{j=1}^{M} \| \varepsilon_k^j\|_2 \stackrel{\eqref{eq: eps}}{\le} \sum_{j=1}^{M}\sqrt{d} \hspace{0.5mm} \tau_k R_k^j. 
\end{equation}
Then, considering the inequalities~\eqref{eq: eps}~and~\eqref{eq: abs_epsk}, and for every $b_k$, we give the following proposition.

 \begin{prop}\label{prop: convergence parameters}
 Let $f(\bw)$ be $\mu$-strongly convex and $L$-smooth, and $f^* : = f(\bw^*)$ be the loss function value of the optimal solution of optimization problem~\eqref{eq: our-optimization}. We define a Lyapunov function as
 \begin{alignat}{3}
 \label{eq: convergence}
 \mathds{V}(\bw_k) &:= f(\bw_k) - f^*  \: \\
 \nonumber
 & + \sum_{i=1}^{k_1} \sum_{h=i}^{k_1} \frac{\zeta_h }{\alpha} \|\bw_{k+1-i}-\bw_{k-i} \|_2^2 + \gamma \sum_{j=1}^{M} \|\varepsilon_k^j \|_{\infty}^2,
\end{alignat}
where $\zeta_h = \zeta, h \in [k_1]$ and $\gamma$ are non-negative constants and $k_1 \le k$. By $0 <\rho<1$, $\beta_i - \beta_{i+1} = \beta_{k_1}, i =1, \ldots, k_1-1$, $a \in(0,1]$, $\alpha = a/L $, and $\gamma \geq d \alpha^2 \left( L + 2\beta_1 + (2\rho \alpha)^{-1}\right)$, $\zeta < M/ 6\tau_{k+1}^2 d k_1$, and 
$$\beta_{k_1} \geq \frac{dL + \frac{d}{2\alpha\rho}}{\frac{M}{3\tau_{k+1}^2\zeta}-2dk_1}.$$ 
Then, Lyapunov function~\eqref{eq: convergence} is non-increasing, i.e. $\mathds{V}(\bw_{k+1}) \le \mathds{V}(\bw_k), k \geq 1$.
 \end{prop}
  \begin{IEEEproof}
  See Appendix~\ref{p:prop: convergence parameters}.
  \end{IEEEproof}

Proposition~\ref{prop: convergence parameters} shows that by proper choice of the Lyapunov function parameters, FL with A-LAQ converges. 

 \begin{figure}[t]
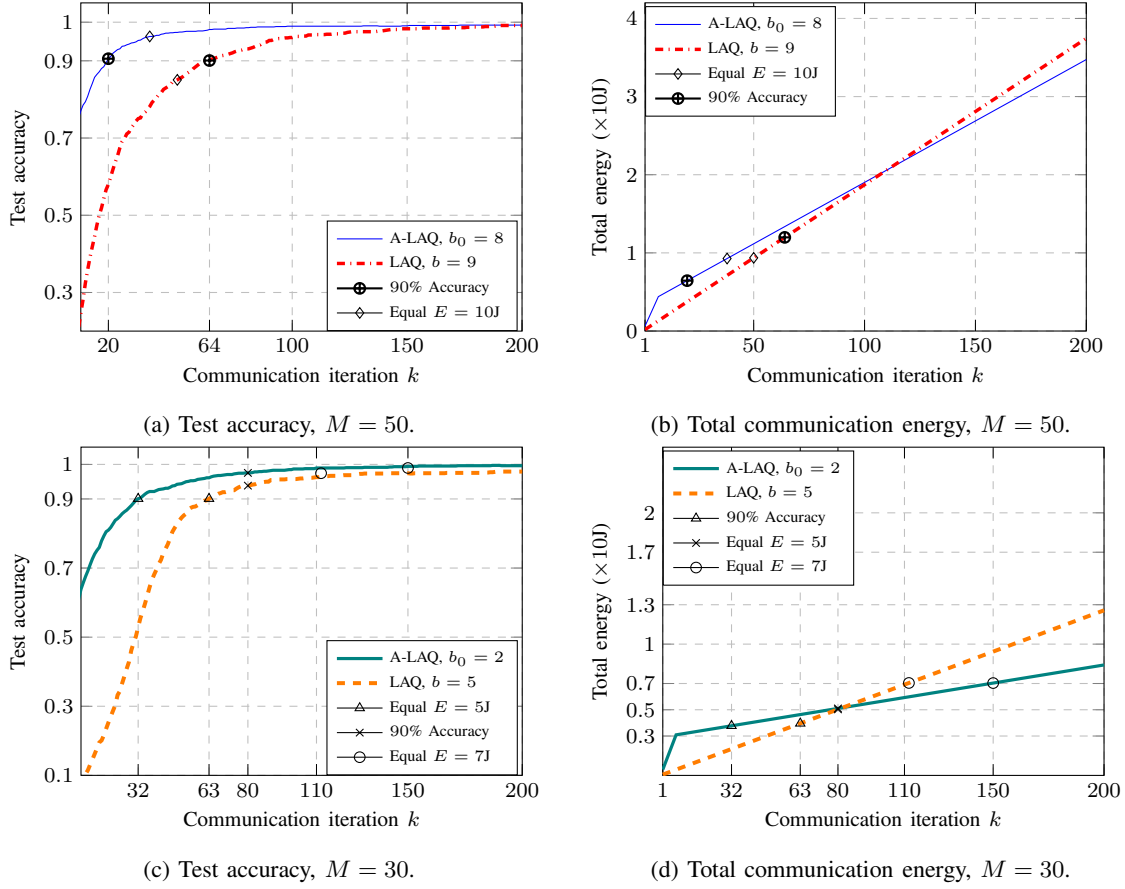

\centering
\begin{minipage}{0.45\columnwidth}
{\scriptsize\input{./figsG/accN50A8L9}}

\subcaption{Test accuracy, $M=50$.}
\label{subfig: accN50A8L9}
\end{minipage}
\hspace{1mm}
\begin{minipage}{0.45\columnwidth}
{\scriptsize\input{./figsG/EN50A8L9}}

\subcaption{Total communication energy, $M=50$.}
\label{subfig: EN50A8L9}
\end{minipage}
\begin{minipage}{0.45\columnwidth}

{\scriptsize\input{./figsG/accN30A2L5}}
\subcaption{Test accuracy, $M = 30$.}
\label{subfig: accN30A2L5}
\end{minipage}
\hspace{1mm}
\begin{minipage}{0.45\columnwidth}

{\scriptsize\input{./figsG/EN30A2L5}}
\subcaption{Total communication energy, $M = 30$.}
\label{subfig: EN30A2L5}
\end{minipage}

\caption{Comparison of A-LAQ and LAQ for a) Test accuracy, and b) Total communication energy for $M = 50$, $b = 9$, $b_0 =8$, $b^{\max}=32$ and $k_0 = 7$. c) Test accuracy, and b) Total communication energy for $M = 30$, $b = 5$, $b_0 =2$, $b^{\max}=32$ with $k_0~=~7$.
}
\label{fig: N50N30}
\end{figure}

\section{Numerical Results}\label{section: numerical}

 In this section, we illustrate our results from the previous sections and numerically show the extensive impact of A-LAQ on FL training. We consider solving a convex regression problem over a wireless network using a real-world dataset. To this end, we extract a binary dataset from MNIST (hand-written digits) by keeping only samples of digits 0 and 1 and then setting their labels to -1 and +1, respectively. We then randomly split the resulting dataset of 12600 samples among $M$ worker nodes, each having $\{(\bx_{ij}, y_{ij})\}$, where  $\bx_{ij} \in \R ^{784}$ is a data sample $i$, which is a vectorized image at node $j \in [M]$ with corresponding digit label $y_{ij} \in \{-1, +1\}$. We use the following training loss function \cite{koh2007interior}
\vspace{-0.01\textheight}
\begin{equation}
    f(\bw) = \sum_{j =1}^M \rho^j \sum_{i =1}^{|D_j|} \frac{1}{|D_j|} \log\left(1 + e^{-\bw^T\bx_{ij} y_{ij}}\right) + \frac{\lambda}{2} \| \bw\|_2^2,
\end{equation}
where $\lambda \in\left(0,1\right)$ is a given regularization parameter and each worker node $j \in [M]$ has the same number of samples, namely $|D_j| = |D_i| = |D|/M, \forall i, j \in [M]$. 

We consider OFDMA for the uplink in a single cell system with the coverage radius of $\ell_c = 1$ Km. There are $L_p$ cellular links on $S_c$ subchannels. We model the subchannel power gain $h_l^s = \phi/(\ell^j)^3$, where $\ell^j$ is the distance between each client to the master node, following the Rayleigh fading, where $\phi$ has an exponential distribution with unitary mean. We consider the noise power in each subchannel as $-170$ dBm/Hz and the maximum transmit power of each link as $23$ dBm. We assume that $S_c=64$ subchannels, the total bandwidth of $10$ MHz, and the subchannel bandwidth of $150$ KHz.
 
 \begin{figure}[t]
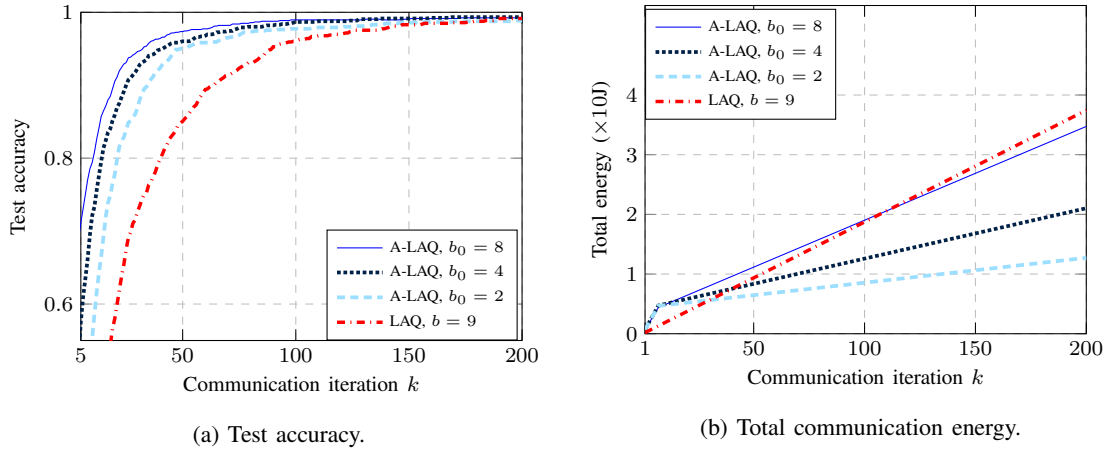

\centering
\begin{minipage}{0.45\columnwidth}
{\scriptsize\input{./figsG/accN50A}}

\subcaption{Test accuracy.}
\label{subfig: accN50A}
\end{minipage}
\hspace{1mm}
\begin{minipage}{0.45\columnwidth}

{\scriptsize\input{./figsG/EN50A}}
\subcaption{Total communication energy.}
\label{subfig: EN50A}
\end{minipage}

\caption{Comparison between LAQ with $b=9$, and A-LAQ with $b_0=2$, $4$ and $8$ for $M=50$. a) Test accuracy shows that all three A-LAQ scenarios outperform LAQ. b) A-LAQ with smaller $b_0$ performs better in an energy limited FL.
}
\label{fig: N50}
\end{figure}

Fig.~\ref{fig: N50N30} illustrates A-LAQ performance and compares it with LAQ. Figs.~\ref{subfig: accN50A8L9}~and~\ref{subfig: EN50A8L9} show test accuracy for $M = 50$, $b = 9$, $b_0 =8$, $b^{\max}=32$ and $k_0 = 7$ is obtained. Each pair of black marks demonstrates the comparison between A-LAQ and LAQ either for the same energy budget $E$ or the same test accuracy. For $E=10$J, we obtain $K = 38$ for A-LAQ with test accuracy of $96$\% , and $K = 50$ for LAQ, with test accuracy of $85$\%. Besides, we observe that for achieving a test accuracy of~$90$\%, A-LAQ spends $50$\% less energy and requires a smaller $K$ than LAQ. 

Figs.~\ref{subfig: accN30A2L5}~and~\ref{subfig: EN30A2L5} address the test accuracy and total spent communication energy for $M = 30$, $b = 5$, $b_0 =2$, $b^{\max}=32$ with $k_0~=~7$. Similar to the previous arguments, for an equal test accuracy of $90$\%, A-LAQ outperforms LAQ by spending approximately the same energy but smaller $K$. For an energy budget $E = 5$J, A-LAQ and LAQ calculate the same $K$, but the test accuracy for A-LAQ is $4$\% higher than LAQ. We also observe that for $k \ge 80$, the total spent communication energy in A-LAQ is lower than LAQ, while the test accuracy of LAQ and A-LAQ are quite similar. Thus, when high communication energy resources are available, A-LAQ requires lower communication energy than LAQ to perform $K$ iterations. 

Fig.~\ref{fig: N50} compares A-LAQ performance of test accuracy and total communication energy for $M=50$, with different values of $b_0 = 8$, $4$, and~$2$. Fig.~\ref{subfig: accN50A} shows test accuracy, and we observe that LAQ has the lowest value of test accuracy for all iterations. Fig.~\ref{subfig: EN50A} demonstrates the total communication energy, which A-LAQ with $b_0 =2$ and $b_0=5$, spends lower energy, while having very close test accuracy to A-LAQ with $b_0=8$. We conclude that A-LAQ with smaller $b_0$ outperforms A-LAQ with higher $b_0$ in terms of energy expenditure and test accuracy for the same energy budget.

\section{Conclusion}\label{section: conclusion}
In this paper, we considered Federated Learning and the LAQ algorithm and proposed an adaptive transmission framework, A-LAQ, by significantly extending LAQ. Different from LAQ, A-LAQ used an adaptive number of communication bits in a communication energy-limited situation. We analyzed the convergence of A-LAQ, and we showed that A-LAQ could achieve a better performance in test accuracy (by an $11$\% increase) while reducing the communication energy by $50$\%.

\emph{Future Work:} Our future work involves extending A-LAQ to communication-efficient scenarios with the best client selection policy. Also, we will consider the computation energy of clients and obtain the optimal sequences of bits to achieve a communication-computation energy-efficient A-LAQ.

\bibliographystyle{./MetaFiles/IEEEtran}
\bibliography{./MetaFiles/References}

\begin{thebibliography}{10}
\providecommand{\url}[1]{#1}
\csname url@samestyle\endcsname
\providecommand{\newblock}{\relax}
\providecommand{\bibinfo}[2]{#2}
\providecommand{\BIBentrySTDinterwordspacing}{\spaceskip=0pt\relax}
\providecommand{\BIBentryALTinterwordstretchfactor}{4}
\providecommand{\BIBentryALTinterwordspacing}{\spaceskip=\fontdimen2\font plus
\BIBentryALTinterwordstretchfactor\fontdimen3\font minus
  \fontdimen4\font\relax}
\providecommand{\BIBforeignlanguage}[2]{{%
\expandafter\ifx\csname l@#1\endcsname\relax
\typeout{** WARNING: IEEEtran.bst: No hyphenation pattern has been}%
\typeout{** loaded for the language `#1'. Using the pattern for}%
\typeout{** the default language instead.}%
\else
\language=\csname l@#1\endcsname
\fi
#2}}
\providecommand{\BIBdecl}{\relax}
\BIBdecl

\bibitem{konevcny2016federated}
J.~Kone{\v{c}}n{\`y}, H.~B. McMahan, F.~X. Yu, P.~Richt{\'a}rik, A.~T. Suresh,
  and D.~Bacon, ``Federated learning: {S}trategies for improving communication
  efficiency,'' \emph{arXiv preprint arXiv:1610.05492}, 2016.

\bibitem{hellstrom2022wireless}
H.~Hellstr{\"o}m, J.~M.~B. da~Silva~Jr, M.~M. Amiri, M.~Chen, V.~Fodor, H.~V.
  Poor, C.~Fischione \emph{et~al.}, ``Wireless for {M}achine {L}earning: {A}
  {S}urvey,'' \emph{Foundations and Trends{\textregistered} in Signal
  Processing}, vol.~15, no.~4, pp. 290--399, 2022.

\bibitem{9264742}
Z.~Yang, M.~Chen, W.~Saad, C.~S. Hong, and M.~Shikh-Bahaei, ``Energy efficient
  {F}ederated {L}earning over wireless communication networks,'' \emph{IEEE
  Transactions on Wireless Communications}, vol.~20, no.~3, pp. 1935--1949,
  2021.

\bibitem{8917592}
S.~Samarakoon, M.~Bennis, W.~Saad, and M.~Debbah, ``Distributed {F}ederated
  {L}earning for ultra-reliable low-latency vehicular communications,''
  \emph{IEEE Transactions on Communications}, vol.~68, no.~2, pp. 1146--1159,
  2020.

\bibitem{mahmoudi2020cost}
A.~Mahmoudi, H.~S. Ghadikolaei, and C.~Fischione, ``Cost-efficient distributed
  optimization in machine learning over wireless networks,'' in \emph{{IEEE}
  International Conference on Communications (ICC)}, 2020.

\bibitem{mahmoudi20spawc}
------, ``Machine learning over networks: Co-design of distributed optimization
  and communications,'' in \emph{{IEEE} International Workshop on Signal
  Processing Advances in Wireless Communications (SPAWC)}, 2020.

\bibitem{mahmoudi2022fedcau}
A.~Mahmoudi, H.~S. Ghadikolaei, J.~M.~B. Da~Silva, and C.~Fischione,
  ``Fed{C}au: {A} proactive stop policy for communication and computation
  efficient {F}ederated {L}earning,'' \emph{arXiv preprint arXiv:2204.07773},
  2022.

\bibitem{amiri2020federated}
M.~M. Amiri, D.~Gunduz, S.~R. Kulkarni, and H.~V. Poor, ``Federated {L}earning
  with quantized global model updates,'' \emph{arXiv preprint
  arXiv:2006.10672}, 2020.

\bibitem{9305988}
N.~Shlezinger, M.~Chen, Y.~C. Eldar, H.~V. Poor, and S.~Cui, ``{UV}e{QF}ed:
  {U}niversal vector quantization for {F}ederated {L}earning,'' \emph{IEEE
  Transactions on Signal Processing}, vol.~69, pp. 500--514, 2021.

\bibitem{8889996}
F.~Sattler, S.~Wiedemann, K.-R. Müller, and W.~Samek, ``Robust and
  communication-efficient {F}ederated {L}earning from non-i.i.d. data,''
  \emph{IEEE Transactions on Neural Networks and Learning Systems}, vol.~31,
  no.~9, pp. 3400--3413, 2020.

\bibitem{9148987}
S.~Li, Q.~Qi, J.~Wang, H.~Sun, Y.~Li, and F.~R. Yu, ``{GGS}: {G}eneral gradient
  sparsification for {F}ederated {L}earning in edge computing,'' in \emph{ICC
  2020 - 2020 IEEE International Conference on Communications (ICC)}, 2020, pp.
  1--7.

\bibitem{9413697}
D.~Jhunjhunwala, A.~Gadhikar, G.~Joshi, and Y.~C. Eldar, ``Adaptive
  quantization of model updates for communication-efficient {F}ederated
  {L}earning,'' in \emph{ICASSP 2021 - 2021 IEEE International Conference on
  Acoustics, Speech and Signal Processing (ICASSP)}, 2021, pp. 3110--3114.

\bibitem{9207963}
H.~Sun, X.~Ma, and R.~Q. Hu, ``Adaptive {F}ederated {L}earning with gradient
  compression in uplink {NOMA},'' \emph{IEEE Transactions on Vehicular
  Technology}, vol.~69, no.~12, pp. 16\,325--16\,329, 2020.

\bibitem{9355797}
P.~Han, S.~Wang, and K.~K. Leung, ``Adaptive gradient sparsification for
  efficient {F}ederated {L}earning: {A}n online learning approach,'' in
  \emph{2020 IEEE 40th International Conference on Distributed Computing
  Systems (ICDCS)}, 2020, pp. 300--310.

\bibitem{9728013}
W.~Yang, Y.~Yang, X.~Dang, H.~Jiang, Y.~Zhang, and W.~Xiang, ``A novel adaptive
  gradient compression approach for communication-efficient {F}ederated
  {L}earning,'' in \emph{2021 China Automation Congress (CAC)}, 2021, pp.
  674--678.

\bibitem{9238427}
J.~Sun, T.~Chen \emph{et~al.}, ``{L}azily {A}ggregated {Q}uantized
  {G}radient~({LAQ}) innovation for communication-efficient federated
  learning,'' \emph{IEEE Transactions on Pattern Analysis and Machine
  Intelligence}, vol.~44, no.~4, pp. 2031--2044, 2022.

\bibitem{9563954}
N.~C. Thompson, K.~Greenewald \emph{et~al.}, ``Deep learning's diminishing
  returns: The cost of improvement is becoming unsustainable,'' \emph{IEEE
  Spectrum}, vol.~58, no.~10, pp. 50--55, 2021.

\bibitem{miller1971minimizing}
B.~L. Miller, ``On minimizing nonseparable functions defined on the integers
  with an inventory application,'' \emph{SIAM Journal on Applied Mathematics},
  vol.~21, no.~1, pp. 166--185, 1971.

\bibitem{koh2007interior}
K.~Koh, S.-J. Kim, and S.~Boyd, ``An interior-point method for large-scale
  $\ell_1$-regularized logistic regression,'' \emph{Journal of Machine Learning
  Research}, vol.~8, no. Jul, pp. 1519--1555, 2007.

\end{thebibliography}

\appendices
\section{}\label{A: proof}
\subsection{Proof of Lemma~\ref{lemma: g}}\label{p: lemma: g}
 This proof is ad-absurdum. Assume that The sequences of $E_f(\bw_k,k; \calI_k^j)$ is not discrete convex. Therefore, there is a $k > 1$ such that $E_f(\bw_k,k; \calI_k^j) > E_f(\bw_{k-1},k-1; \calI_{k-1}^j)$ and $E_f(\bw_k,k; \calI_k^j) > E_f(\bw_{k+1},k+1; \calI_{k+1}^j)$. According to the statement of Lemma~\ref{lemma: g}, since $b_k = b^{\max}$, $k\le k_0$, we consider $\sum_{k'=1}^k E_{k'} = k E_1$. Besides, $f(\bw)$ is $\mu$-strongly convex and $L$-smooth, which means the sequence of $f(\bw_k)$ have the descent behavior w.r.t. $k$, and satisfies $f(\bw_k)-f(\bw_{k+1}) \le f(\bw_{k-1}-f(\bw_k)$. According to the definition of $E_f(\bw_k,k; \calI_k^j) = k E_1/(f(\bw_0)-f(\bw_{k}))$, we have $f(\bw_0)-f(\bw_{k}) = f(\bw_0)-f(\bw_{k-1}) +f(\bw_{k-1}) -f(\bw_{k}) \ge f(\bw_0)-f(\bw_{k-1})$, which means that both numerator and denominator of $E_f(\bw_{k+1},k+1; \calI_{k+1}^j)$ are non-decreasing w.r.t. $k$. Now, if we assume that $E_f(\bw_k,k; \calI_k^j) > E_f(\bw_{k-1},k-1; \calI_{k-1}^j)$ and $E_f(\bw_k,k; \calI_k^j) > E_f(\bw_{k+1},k+1; \calI_{k+1}^j)$, it results in a decrease in the denominator from $k$ to $k+1$, thus we obtain that $f(\bw_0)-f(\bw_{k}) \ge f(\bw_0)-f(\bw_{k+1})$ which is in contradiction with the behavior of $f(\bw_k)$. Therefore, we conclude that $E_f(\bw_k,k; \calI_k^j)$ is discretely convex.


\subsection{Proof of Proposition~\ref{prop: k_0}}\label{p: prop: k_0}
First consider that $k_0=k_e$, it means that the energy budget is determining $k_0$. As we mentioned in~\ref{p: lemma: g}, $E_k = E_0$ and $\sum_{k'=1}^k E_{k'} = k E_0$. Thus, when $E_0 > E- kE_0 $, it results in energy limitation and then $k_0 = K = k$.

Next, consider that $k_0=k_f$, according to~Lemma~\ref{lemma: g}, $ E_f(\bw_k,k; \calI_k^j)$ is discrete convex and we obtain $k_0 = k$ when~$ E_f(\bw_k,k; \calI_k^j) - E_f(\bw_{k-1},k-1; \calI_{k-1}^j) > 0 $, see~\cite{mahmoudi2020cost}. Thus, 
\begin{alignat}{3}
&E_f(\bw_k,k; \calI_k^j) - E_f(\bw_{k-1},k-1; \calI_{k-1}^j) =  \: \\
\nonumber
&\frac{kE_0}{f(\bw_0)-f(\bw_k)} - \frac{(k-1)E_0}{f(\bw_0)-f(\bw_{k-1})} =  \: \\
\nonumber
&\frac{kE_0}{f(\bw_0)-f(\bw_k)} - \frac{(k-1)E_0}{f(\bw_0)-f(\bw_{k-1})} > 0,\: \\
\nonumber
&k < \frac{f(\bw_0)-f(\bw_k)}{f(\bw_{k-1})-f(\bw_{k})}.
\end{alignat}
Therefore, the proof is complete. 

\subsection{Proof of Lemma~\ref{lemma: EK}}\label{p: lemma: EK}
This proof is similar to~\ref{p: prop: k_0}, when $k_0 = k_f$, but with considering adaptive $b_k$. Since at each iteration $k$, we compute $b_{k+1} = \lceil \eta_k b_k\rceil$, the possible causal way to obtain $K$ is to use the current information of communication energy $E_k/b_k$. Thus, we obtain $K$ when the causal approximation of $E_{k+1}$, i.e., $b_{k+1} E_k/b_k$ is greater than $E - \sum_{k'=1}^{k} E_{k'}$. Thus, we obtain the inequality~\eqref{eq: EK}. 

\subsection{Proof of Proposition~\ref{prop: convergence parameters}}\label{p:prop: convergence parameters}
 According to~\cite{9238427},
\begin{alignat}{3}\label{eq: Eq21}
 \|\varepsilon_{k+1}^j \|_\infty^2& \le \tau^2 (R_{k+1}^{j})^2 \: \\
 \nonumber
 & \le 3\tau^2 L_j \|\bw_{k+1}-\bw_k\|_2^2 + 3\tau^2\|\varepsilon_{k}^j \|_\infty^2,
\end{alignat}
where $\|\varepsilon_{k}^j \|_\infty^2 \le \tau^2(R_{k}^{j})^2$, 
\begin{equation}\label{eq: fixed_tau0}
    \tau^2 (R_{k+1}^{j})^2 \le 3\tau^2 L_j \|\bw_{k+1}-\bw_k\|_2^2 + 3\tau^4(R_{k}^{j})^2.
\end{equation}
{According to~\eqref{eq: Eq21}, we derive} the following inequality for A-LAQ.
\begin{equation}\label{eq: tau_k0}
    \tau_{k+1}^2 (R_{k+1}^{j})^2 \le 3\tau_{k+1}^2 L_j \|\bw_{k+1}-\bw_k\|_2^2 + 3\tau_{k+1}^2 \tau_k^2(R_{k}^{j})^2.
\end{equation}

By inserting~\eqref{eq: tau_k0} into~\eqref{eq: convergence} , we obtain the one-step Lyapunov function as 

\begin{alignat}{3}
\nonumber
\mathds{V}(\bw_{k+1})  - \mathds{V}(\bw_k) & \le-\alpha \langle \nabla f(\bw_k),\bq_k\rangle + \frac{\alpha}{2} \|\nabla f(\bw_k)\|_2^2\: \\
 \nonumber
 &  + (\frac{L}{2} + \beta_1 + 3\gamma \tau_{k+1}^2L_j^2) \|\bw_{k+1}-\bw_k \|_2^2  \: \\
 \nonumber
 & + \sum_{i=1}^{k_1-1} (\beta_{i+1}-\beta_i) \|\bw_{k+1-i}-\bw_{k-i} \|_2^2 \: \\
 \nonumber
 & - \beta_{k_1}\|\bw_{k+1-k_1}-\bw_{k-k_1} \|_2^2\: \\
 \nonumber
 & + \gamma(3\tau_{k+1}^2 -1)\sum_{j=1}^{M} \|\varepsilon_k^j \|_{\infty}^2\: \\
 \label{eq: convergence0}
 & + 3\gamma \tau_{k+1}^2 \sum_{j=1}^M \|\bq_{k-1}^j-\bq_k^j \|_2^2.
\end{alignat}
By replacing $\bq_k = \nabla f(\bw_k)-\varepsilon_k$, $\bw_{k+1}-\bw_k = \alpha \bq_k$, and for any $\rho > 0$ 
\begin{equation}\label{eq: inner product}
    \langle \nabla f(\bw_k),\varepsilon_k \rangle \le \frac{\rho}{2} \|\nabla f(\bw_k)\|_2^2 + \frac{1}{2\rho} \|\varepsilon_k \|_{2}^2,
\end{equation}
and defining $A_{k+1} := L + 2\beta_1 + 6\gamma \tau_{k+1}^2L_j^2$, we simplify~\eqref{eq: convergence0} as
\begin{alignat}{3}
\nonumber
\mathds{V}(\bw_{k+1}) &- \mathds{V}(\bw_k)  \le  \|\nabla f(\bw_k)\|_2^2 {\color{black}{\left(\alpha^2 A_{k+1}  - \frac{\alpha}{2} + \frac{\alpha \rho}{2}\right)}}\: \\
 \nonumber
 & + \| \varepsilon_k\|_2^2 \left(\alpha^2 A_{k+1}   + \frac{\alpha }{2\rho}\right)   \: \\
 \nonumber
 &+ {\color{black}{\left( \frac{3\gamma \tau_{k+1}^2 \zeta_{k_1}}{\alpha^2 M} - \beta_{k_1}\right)} }\|\bw_{k+1-k_1}-\bw_{k-k_1} \|_2^2\: \\
 \nonumber
 & + {\color{black}{\sum_{i=1}^{k_1-1} \left(\beta_{i+1}-\beta_i +\frac{3\gamma \tau_{k+1}^2 \zeta_{i}}{\alpha^2 M}\right)}} \|\bw_{k+1-i} \: \\
 \nonumber
 &  -\bw_{k-i} \|_2^2 + \gamma \left(3\tau_{k+1}^2 -1\right)\sum_{j=1}^{M} \|\varepsilon_k^j \|_{\infty}^2 \: \\
 \nonumber
 & \le  \|\nabla f(\bw_k)\|_2^2 {\color{black}{\left(\alpha^2 A_{k+1}  - \frac{\alpha}{2} + \frac{\alpha \rho}{2}\right)}}\: \\
 \nonumber
 & + {\color{black}{\left( \frac{3\gamma \tau_{k+1}^2 \zeta_{k_1}}{\alpha^2 M} - \beta_{k_1}\right)}} \|\bw_{k+1-k_1}-\bw_{k-k_1} \|_2^2\: \\
 \nonumber
 & + {\color{black}{\sum_{i=1}^{k_1-1} \left(\beta_{i+1}-\beta_i +\frac{3\gamma \tau_{k+1}^2 \zeta_{i}}{\alpha^2 M}\right)}} \|\bw_{k+1-i} -\: \\
 \nonumber
 &  \bw_{k-i} \|_2^2 + {\color{black}{\left(d \alpha^2 A_{k+1} + \frac{d \alpha }{2\rho}+ \gamma \left(3\tau_{k+1}^2 -1\right)\right)}} \times \: \\
\label{eq: convergence1}
 &  \left[ \sum_{j=1}^{M} \|\varepsilon_k^j \|_{\infty}\right]^2.
\end{alignat}

Then, by setting the coefficient to be non-positive, we complete the proof.

\end{document}
\subsection{Proof of Proposition~\ref{prop: descent}}\label{p:prop: descent}
First, we consider the following inequality~\cite{9238427} 
\begin{alignat}{3} \label{original}
&f(\bw_{k+1})-f(\bw_k)\le - \frac{\alpha}{2} \| \nabla f(\bw_k)\|_2^2 + \alpha\|\varepsilon_k\|_2^2\: \\
\nonumber
&= {\alpha}\left( \frac{\|\nabla f(\bw_k)\|_2}{\sqrt{2}}+\|\varepsilon_k\|_2\right)\left( \frac{-\|\nabla f(\bw_k)\|_2}{\sqrt{2}}+\|\varepsilon_k\|_2\right).
\end{alignat}
Now, we investigate that under which conditions we obtain ${-\|\nabla f(\bw_k)\|_2}/{\sqrt{2}}+\|\varepsilon_k\|_2 \le 0 $, which results in $f(\bw_{k+1})~-~f(\bw_k)~\le~0$. According to~\eqref{eq: eps}~and~\eqref{eq: abs_epsk}, $\|\varepsilon_k\|_2  \le \sum_{j=1}^{M}\sqrt{d} \hspace{0.5mm} \tau_k R_k^j$, and we replace $R_k^j =\| \nabla f^j(\bw_k) - \bq^j({\bw}_{k-1})\|_\infty = \| \nabla f^j(\bw_k) - \bq^j({\bw}_{k})+\delta \bq_k^j\|_\infty $, thus
\begin{alignat}{3}\label{o1}
&\|\varepsilon_k\|_2 -\frac{\|\nabla f(\bw_k)\|_2}{\sqrt{2}}\le \: \\
\nonumber
& \tau_k\sum_{j=1}^{M} \|\nabla f^j(\bw_k) - \bq^j({\bw}_{k})+\delta \bq_k^j\|_\infty-\frac{\|\nabla f(\bw_k)\|_2}{\sqrt{2}}
\end{alignat}